%% file: main.tex
\documentclass{article}
\usepackage{ijcai19}

\usepackage{times}
\usepackage{soul}
\usepackage{url}
\usepackage[hidelinks]{hyperref}
\usepackage[utf8]{inputenc}
\usepackage[small]{caption}
\usepackage{graphicx}
\usepackage{amsmath}
\usepackage{booktabs}
\usepackage{algorithm}
\usepackage{algorithmic}
\usepackage{comment}

\newcommand{\mat}[1]{\mathbf{#1}}
\newcommand\mydots{\makebox[1em][c]{.\hfil.\hfil.}}
\usepackage{amssymb,amsfonts}
\usepackage{multirow}
\usepackage{subfig}
\usepackage{dblfloatfix}

\title{SparseSense: Human Activity Recognition from Highly Sparse Sensor Data-streams Using Set-based Neural Networks}

\author{
	Alireza Abedin, 
	S. Hamid Rezatofighi, 
	Qinfeng Shi,  
	Damith C. Ranasinghe
	\affiliations
	School of Computer Science, The University of Adelaide, Australia\\
	\emails
	\{alireza.abedinvaramin, hamid.rezatofighi, javen.shi, damith.ranasinghe\}@adelaide.edu.au
}

\begin{document}
\maketitle



%
%
%

\begin{abstract}
Batteryless or so called \textit{passive} wearables are providing new and innovative methods for human activity recognition (HAR), especially in healthcare applications for older people. Passive sensors are low cost, lightweight, unobtrusive and desirably disposable; attractive attributes for healthcare applications in hospitals and nursing homes. Despite the compelling propositions for sensing applications, the data streams from these sensors are characterised by \textit{high sparsity}---the time intervals between sensor readings are irregular while the number of readings per unit time are often limited. 
In this paper, we rigorously explore the problem of learning activity recognition models from temporally sparse data. 
We describe how to learn directly from sparse data using a deep learning paradigm in an end-to-end manner. We demonstrate significant classification performance improvements on real-world passive sensor datasets from older people over the state-of-the-art deep learning human activity recognition models. Further, we provide insights into the model's behaviour through complementary experiments on a benchmark dataset and visualisation of the learned activity feature spaces. 

\end{abstract}


\section{Introduction}


\begin{figure*}[t]
	\centering
	\includegraphics[width=\textwidth, keepaspectratio]{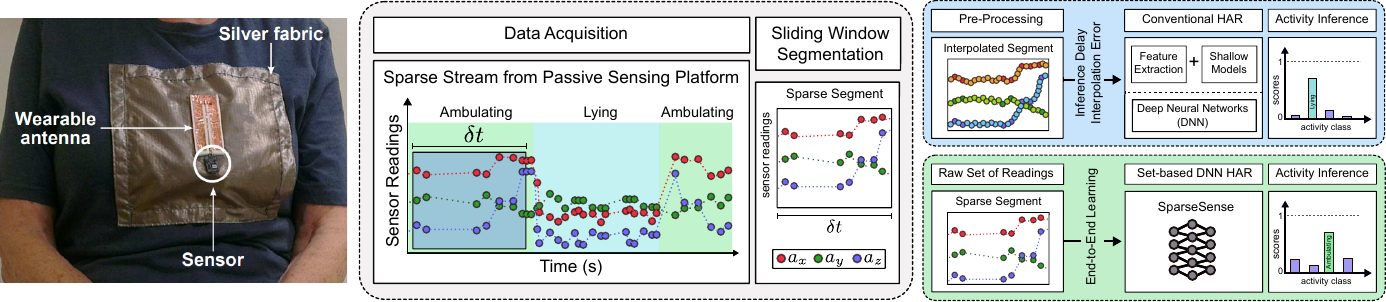}
	\caption{\textbf{Left}:~Older volunteer wearing a passive sensor over their clothing in the clinical rooms public datasets used in this work (datasets and figure from~\protect\cite{clinical}). \textbf{Right}:~An overview of the conventional sparse data-stream classification pipeline (blue plane) versus our novel set-based deep learning pipeline (green plane). Initially, data-streams from passive sensors are partitioned. The conventional pipeline then applies interpolation pre-processing on the sparse segments to synthesise fixed temporal context for model training and inference. In contrast, our proposed approach elegantly allows end-to-end learning of activity recognition models directly from sparse segments to deliver highly accurate classification decisions.}
	\label{fig:workflow}
\end{figure*}

Understanding human activities using wearables is the basis for an increasing number of healthcare applications such as rehabilitation, gait analysis, falls detection and falls prevention~\cite{bulling}. In particular, older people have expressed a preference for unobtrusive and wearable sensing modalities~\cite{Govercin2010UserReqFallDetect,roberto2018ponealarms}. While traditional wearables employ battery powered devices, new opportunities for human activity recognition applications, especially in healthcare, are being created by batteryless or \textit{passive} wearables operating on harvested energy~\cite{chen2015paired,lemey2016wearable}. In contrast to using often bulky and obtrusive battery powered wearables, passive sensing modalities provide maintenance-free, often disposable, unobtrusive and lightweight devices highly desirable to both older people and healthcare providers. However, the very nature of these sensors leads to new challenges. 
\vspace{1mm}

\noindent{\textbf{Problem.}~}The process of operating a batteryless sensor and transmitting the data captured is reliant on harvested power. Due to variable times to harvest adequate energy to operate sensors, the data-streams generated are highly sparse with variable inter-sample times. We illustrate the problem in Fig.~\ref{fig:workflow} for a data stream captured by a body-worn passive sensor. We can see two key artefacts: \textit{i}) the variable time intervals between sensor data reporting times; and \textit{ii}) the relatively low average sampling rate. 
In this paper, we consider the problem of learning human activity recognition (HAR) models from \textit{sparse data-streams} using a deep learning paradigm in an \textit{end-to-end} manner.



\vspace{1mm}



\noindent\textbf{Current Approaches.~}Wearable sensors generate time-series data. Consequently, the dominant human activity recognition pipeline uses fixed duration sliding window partitioning  
to feed neural networks during both training and inference stages~\cite{survey,ensemble,deepconvlstm,bilstm,multichannelcnn,convmobile}. 
When dealing with sparse data partitions, a common remedy is to rely on interpolation techniques as a pre-processing step to synthesise sensor observations to obtain a fixed size representation from time-series partitions as illustrated in Fig.~\ref{fig:workflow}~\cite{interpolate,Gu2018LocomotionPhones}. 
However, we recognize two key issues with an interpolated sparse data-stream:
\begin{itemize}
    \item Interpolating between sensor readings that are temporarily distant can potentially lead to poor approximations of missing measurements and contextual activity information. 
    Accordingly, adoption of convolutional filters or recurrent layers to extract temporal patterns from the poorly approximated measurements may potentially propagate the estimation errors to the activity recognition model---we substantiate this through extensive experiments in Section \ref{results}. 
    \item Interpolation is as an intermediate processing step that prevents end-to-end learning of activity recognition models directly from raw data and introduces real-time prediction delays in time critical applications---we demonstrate the time overheads imposed on inference in Section~\ref{results}.
\end{itemize}
\vspace{1mm}

\noindent\textbf{Our Approach.~} 
Instead of relying on the naturally poor temporal correlations between  consecutively  received samples  in sparse data-streams, we consider incentivizing  the  activity  recognition  model  to uncover discriminative representations from the input sensory data partitions of various sizes to distinguish different activity categories. Our intuition is that a few information bearing sensor samples, although not temporally consistent, can capture adequate amount of information. Therefore, we propose learning HAR models directly from sparse data-streams. An illustrative summary of our proposed methodology for sparse data-stream classification in comparison with the conventional treatment is presented in Fig. \ref{fig:workflow}. 
%



In this paper, we describe how human activity recognition with sparse data-streams can be elegantly handled using deep neural networks in an end-to-end learning process. Given that we no longer rely on often poor temporal information, we represent sparse data stream partitions as unordered sets with various cardinalities from which embeddings capable of discriminating activities can be learned. Our approach is inspired by recent research efforts to investigate set-based deep learning paradigms to address a new family of problems where inputs \cite{pointnet,deepsets} of the task are naturally expressed as sets with unknown and unfixed cardinalities.
Therefore, our approach here is to develop activity recognition models that can learn and predict from incomplete sets of sensor observations, without requiring any extra interpolation efforts.  
\vspace{1mm}

\noindent\textbf{Contribution.~}In particular: \textit{\textbf{i)}}~We solve a new problem with a deep neural network formulation---learning from sparse sensor data-streams in an end-to-end manner; \textit{\textbf{ii)}}~We show that set learning can tolerate missing information which otherwise would not be possible with conventional DNN; and \textit{\textbf{iii)}}~We demonstrate that our novel treatment of the problem yields significantly outperforming recognition models with lower inference delays compared with the state-of-the-art on naturally sparse public datasets---over 4\% improvement in the best case. We further compare with a benchmark HAR dataset and provide deeper insights into the performance improvements obtained from our proposed approach.










\input{Methodology}

\input{Experiments}

\input{Conclusions}


\newpage
\bibliographystyle{named}
\bibliography{main}

\end{document}

%% file: Methodology.tex
\section{Methodology}
We first present a formal description of human activity recognition problem with sparse data-streams and introduce the notations used throughout this paper before elaborating on our proposed activity recognition framework to learn directly from sparse data-streams in an end-to-end manner. 

\subsection{Problem Formulation}

\begin{figure*}[ht]
	\centering
	\includegraphics[width=\textwidth, keepaspectratio]{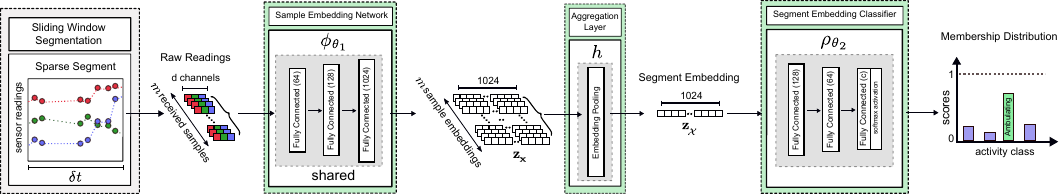}
	\caption{\textit{SparseSense} architecture. The proposed network consumes sets of raw sensor observations with potentially varying cardinalities, uncovers latent projections for individual samples, aggregates sample embeddings into a global segment embedding, and maps the acquired segment embedding to its corresponding activity category. The number of neurons constituting the layers are outlined in parenthesis. All layers utilize ReLUs for non-linear transformation of activations except for the last layer which leverages a softmax activation function.}
	\label{fig:network}
\end{figure*}

Consider a collected data-stream of raw time-series samples from body-worn sensors of the form 
$\mat{S}=(\mat{x}_1,\mat{x}_2,\mydots, \mat{x}_\mathrm{T})$, where $\mat{x}_t\in\mathbb{R}^{\mathrm{d}}$ is a multi-dimensional vector that contains sample measurements over $\mathrm{d}$ distinct sensor channels at time step $t$ and $\mathrm{T}$ is the total length of the sequence. Without loss of generality, we assume a hardware-specific sampling rate for the wearable sensors, denoted by $\mathrm{f}$. 
\subsubsection{HAR with Uniform Time-series Data}

In an ideally controlled laboratory setup, sensor samples are constantly taken at regular intervals of $\frac{1}{\mathrm{f}}$ seconds. In such case, applying the commonly adopted time-series segmentation technique with a sliding window of fixed temporal context $\delta t$ yields the labeled dataset
\begin{equation}
\mathcal{D}_{\textrm{uniform}}=\{(\mat{X}_1, \mat{y}_1),(\mat{X}_2, \mat{y}_2),\mydots, (\mat{X}_n, \mat{y}_n)\},
\label{eq:dataset}
\end{equation}
where $\mat{X}_i=[\mat{x}_i,\mydots,\mat{x}_{i+\mathrm{m}-1}]\in\mathbb{R}^{\mathrm{d}\times \mathrm{m}}$ is a \textit{fixed size} segment of captured sensor readings, $\mathrm{m}=\mathrm{f}\delta t$ is the constant number of received samples, and $\mat{y}_i$ denotes the corresponding one-hot encoded ground-truth from the pre-defined activity space $\mathcal{A}=\{a_1,\mydots,a_c\}$. The acquired dataset can then be utilized to train activity recognition models using out-of-the-box machine learning techniques. 

\subsubsection{HAR with Sparse Time-series Data}
Unfortunately, sparse time-series data often found in real-world deployment settings, especially with passive sensors have variable inter-sensor observation intervals. 
In this case, utilising a fixed time sliding window approach to segment the sparse data-stream results in the labeled dataset:

\begin{equation}
\mathcal{D}_{\textrm{sparse}}=\{(\mathcal{X}_1^{m_1}, \mat{y}_1),(\mathcal{X}_2^{m_2}, \mat{y}_2),\mydots, (\mathcal{X}_n^{m_n}, \mat{y}_n)\},
\label{eq:dataset_sparse}
\end{equation}
where $\mathcal{X}_{i}^{m_i}=\{\mat{x}_i,\mydots,\mat{x}_{i+m_i-1}\}\in{\overbrace{\mathbb{R}^\mathrm{d}\times\mydots\times\mathbb{R}^\mathrm{d}}^{m_i}}$ is a set of sparse sensor observations during a timed window, $m_i\in~\mathbb{N}$ is the cardinality of the obtained observation set, and $\mat{y}_i$ denotes the corresponding activity class. We emphasise that the number of received sensor readings in the time interval $\delta t$ is \textit{unfixed} for different sensory segments and upper bounded by the sensor sampling rate; \textit{i.e.}, for any given sensory segment $\mathcal{X}_{i}^{m_i}$, we have $m_i\leq \mathrm{f}\delta t$.

In this paper, having acquired the training dataset of sparse sensory segments $\mathcal{D}_{\textrm{sparse}}=\{(\mathcal{X}_i^{m_i}, \mat{y}_i)\}_{i=1}^n$, we intend to directly learn a mapping function $\mathcal{F}_{\Theta^*}:2^{\mathbb{R}^{\mathrm{d}}}\to\mathcal{A}$, that operates on input sensory sets with unfixed cardinalities and accurately predicts the underlying activity classes, 
\begin{equation}
\mat{y}_i=\mathcal{F}_{\Theta^*}(\mathcal{X}_i^{m_i})=\mathcal{F}_{\Theta^*}(\{\mat{x}_i,\mydots,\mat{x}_{i+m_i-1}\}), ~~\forall i \in \{1,\mydots,n\}.
\label{eq:ideal mapping function}
\end{equation}
\subsection{SparseSense Framework}

\begin{table*}[!t]
	\centering
	\caption{A comparison of the best performing activity recognition models for the naturally sparse clinical room datasets. Numbers and design choices for the baselines with asterisks are quoted from \protect\cite{interpolate}. The remaining baselines are replicated following their original paper descriptions. For a fair comparison, the reported results are for the best performing configuration of interpolants and window durations for each baseline.} 
	\label{tab:compare}
	\resizebox{!}{26mm}{
		\begin{tabular}{c|c |c |c| c| c c c}
			\textbf{Dataset} & \textbf{HAR Model} & \textbf{Interpolant} & \textbf{Input} & \textbf{Window Size}   & $\textbf{Precision}_{m}$ & $\textbf{Recall}_{m}$  & $\textbf{F-score}_{m}$\\ 
			(clinical room)  &                   &       (acceleration)            &                      &   ($\delta t$)         &   (mean$\pm$std)   &   (mean$\pm$std)   & (mean$\pm$std)   \\ \hline
			\multirow{7}{*}{\textit{Roomset1}} 
			& ${\textrm{SVM}^{lin*}}$    &  Cubic         &   Hand-crafted features   & 4 seconds   &   87.87$\pm$2.55    &   83.44$\pm$1.72   & 84.96$\pm$1.23\\ 
			
			& ${\textrm{SVM}^{rbf*}}$    &  None           &   Hand-crafted features   & 8 seconds   &   90.39$\pm$2.70    &   87.42$\pm$1.42   & 88.45$\pm$1.68\\ 
			
			& CRF$^*$                    &  Linear        &   Hand-crafted features    & 2 seconds   &   85.97$\pm$2.43    &   82.35$\pm$3.08   & 83.73$\pm$2.40\\ 
			
			& Bi-LSTM                    &  Linear        &   Raw sensor readings    & 2 seconds   &   89.97$\pm$0.78    &   85.11$\pm$0.99   & 86.96$\pm$1.06\\ 
			
			& DeepCNN                  &  Quadratic        &   Raw sensor readings    & 4 seconds   &   92.43$\pm$1.21    &   87.93$\pm$1.74   & 89.73$\pm$1.55\\ 
			
			& DeepConvLSTM           &  Linear        &   Raw sensor readings    & 4 seconds   &   91.87$\pm$1.43    &   88.88$\pm$1.79   & 90.42$\pm$1.54\\ 
			
			& \textbf{(Ours) SparseSense}                    &  None           &   Raw sensor readings           & 2 seconds   &   \textbf{95.0$\pm$0.75}    &   \textbf{94.08$\pm$0.78}   & \textbf{94.51$\pm$0.62}\\ 
			
			\hline \\\hline
			\multirow{7}{*}{\begin{tabular}[c]{@{}c@{}}\textit{Roomset2}\end{tabular}} 
			& ${\textrm{SVM}^{lin*}}$    &  Cubic         &   Hand-crafted features   & 2 seconds    &   87.06$\pm$4.10   &   84.00$\pm$2.90   & 84.97$\pm$3.74\\ 
			
			& ${\textrm{SVM}^{rbf*}}$    &  None           &   Hand-crafted features   & 8 seconds    &   90.97$\pm$4.11    &   83.88$\pm$2.04   & 85.53$\pm$2.86\\ 
			
			& CRF$^*$                     &  None          &   Hand-crafted features    & 16 seconds   &   83.68$\pm$6.50    &   78.29$\pm$3.58   & 79.99$\pm$4.76\\ 
			
			& Bi-LSTM                 &  Previous        &   Raw sensor readings    & 2 seconds   &   92.38$\pm$0.91    &   91.4$\pm$0.62   & 91.78$\pm$0.58\\ 
			
			& DeepCNN                    &  Linear        &   Raw sensor readings    & 4 seconds   &   93.11$\pm$0.94    &   91.7$\pm$1.18   & 92.36$\pm$0.99\\ 
			
			& DeepConvLSTM                   &  Previous        &   Raw sensor readings    & 4 seconds   &   94.16$\pm$0.52    &   93.05$\pm$0.78   & 93.77$\pm$0.63\\ 
			
			& \textbf{(Ours) SparseSense}                    &  None           &   Raw sensor readings             & 2 seconds    &   \textbf{97.07$\pm$0.52}    &   \textbf{96.88$\pm$0.34}   & \textbf{96.97$\pm$0.37}\\ \hline
		\end{tabular}}
	\end{table*}

Our work is built upon the insight that incorporating interpolation techniques to recover the missing measurements across large temporal gaps between received sensor observations in sparse data-streams leads to poor estimations and therefore, significant interpolation errors.  As we demonstrate in Section~\ref{results}, the adoption of convolutional filters or recurrent layers to extract temporal patterns from the poorly approximated measurements can potentially propagate the estimation errors to the activity recognition model. 




Instead of forcing the network to exploit the potentially weak temporal correlations in sparse data-streams, 
we propose learning global embeddings from sets that encode aggregated information related to an activity. Therefore, we propose formulating sparse segments as unordered sets with unfixed and unknown number of sensor readings. 
Hence, we design \textit{SparseSense} as a set-based activity recognition framework for the HAR task that directly manipulates sets of received sensor readings with irregular inter-sample observation intervals and outputs the corresponding activity membership distributions. Our approach provides a complete end-to-end learning method that incentivizes the activity recognition model to uncover globally discriminative representations for the input sparse segments with variable number of samples, and distinguish different activity categories accordingly.
\vspace{1mm}

\noindent\textbf{Network Architecture.~}\label{net explain}
The overall architecture of our proposed \textit{SparseSense} network is illustrated in Fig. \ref{fig:network}. Essentially, we approximate the optimal mapping function $\mathcal{F}_{\Theta^*}$ in Eq.~(\ref{eq:ideal mapping function}) through training of a deep neural network  parameterized by ${\Theta}$. The primary task for integrating set learning into deep neural networks is employing a \textit{shared network} to map each set element independently into a higher dimensional embedding space (to facilitate class separability) and adopting a symmetric operation across the element embeddings to generate a global representation for the entire set that does not rely on the set element orderings. We incorporate this pipeline into the building blocks of our network as elucidated in what follows:
 


\textit{Input.~}Adopting sliding window segmentation over the sparse data-stream yields sets of sparsely received sensor observations $\mathcal{X}$ in the pre-defined temporal window $\delta t$, with potentially varying cardinalities. 

\textit{The shared sample embedding network.~}The embedding network $\phi_{\theta_1}:~\mathbb{R}^{\mathrm{d}}\to\mathbb{R}^{\mathrm{z}}$ parameterized by $\theta_1$, operates identically and independently on each sample measurement $\mat{x}$ within the received observation set $\mathcal{X}$ and learns a corresponding higher dimensional projection $\mat{z}_{\mat{x}}\in\mathbb{R}^{\mathrm{z}}$ to alleviate separability of activity features in the new embedding space; \textit{i.e.}, $\mat{z}_{\mat{x}} = \phi_{\theta_1}(\mat{x}) , \forall\mat{x}\in\mathcal{X}$. Technically, $\phi_{\theta_1}$ is a standard multi-layer perceptron (MLP) whose parameters are \textit{shared} between the sensor sample readings; \textit{i.e.}, all samples undergo the same layer operations and are therefore processed identically through a copy of the MLP. 

   

\textit{The aggregation layer.~} Described by $h:\mathbb{R}^{\mathrm{z}}\times\mydots\times\mathbb{R}^{\mathrm{z}}\to\mathbb{R}^{\mathrm{z}}$, the aggregation layer applies a symmetric operation across the latent representations of individual sensor samples and extracts a fixed size global embedding $\mat{z}_{\mathcal{X}}\in\mathbb{R}^{\mathrm{z}}$ to represent the sensory segment as a whole. Thus, for a given sensory segment $\mathcal{X}_i$, we have
\begin{equation}
\mat{z}_{\mathcal{X}_i}=h(\{\mat{z}_{\mat{x}_i},\mydots,\mat{z}_{\mat{x}_{i+m_i-1}}\}). ~~ 
\label{eq:pooling}
\end{equation}
Notably, the shared sample embedding network coupled with the symmetric aggregation layer allow summarizing sparse segments with effective high-dimensional projections that \textit{i}) do not rely on the weak temporal ordering of the sparse samples, and, \textit{ii}) ensure fixed size tensor representations independent of the number of received readings. Inspired by \cite{pointnet}, in this paper, we set $h$ to incorporate a feature-wise maximum pooling across sample embeddings which promises robustness against set element perturbations.  


\textit{The segment embedding classifier.~}Described by $\rho_{\theta_2}:~\mathbb{R}^{\mathrm{z}}\to\mathcal{A}$  parameterized by $\theta_2$ is trained to exploit the segment embeddings $\mat{z}_{\mathcal{X}}$ through multiple layers of non-linearity and predict the corresponding activity class probability distributions $\hat{\mat{y}}$; \textit{i.e.}, $\hat{\mat{y}}= \rho_{\theta_2}(\mat{z}_{\mathcal{X}})$. 
Here, a softmax activation function governs the output of our network to yield posterior probability distributions over the activity space $\mathcal{A}$.
\vspace{2mm}

\textit{Summary.~} Now, we can express the mathematical operations constituting the forward pass of our proposed activity recognition model for a given sparse sensory segment $\mathcal{X}_i$ as:
\begin{equation}
\mathcal{F}_{\Theta}(\mathcal{X}_i^{m_i})=\rho_{\theta_2}\left( h(\{\phi_{\theta_1}(\mat{x}_i),\mydots,\phi_{\theta_1}(\mat{x}_{i+m_i-1})\})\right),
\label{eq:forward}
\end{equation}
where $\Theta$ denotes the collection of all network parameters; \textit{i.e.}, $\Theta=(\theta_1,\theta_2)$. \\

\noindent\textbf{Network Training and Activity Inference.~}
During the training process, the goal is to learn the network parameters $\Theta$ such that the disagreement between the network outputs and the corresponding ground-truth activities is minimised for the training dataset. We can precisely express this discrepancy minimisation by adopting an end-to-end optimisation of the negative log-likelihood loss function ${\mathcal{L}_\textrm{NLL}}$ on the training dataset $\mathcal{D}_{\textrm{sparse}}$; \textit{i.e.},
 \begin{equation}
\Theta^*=\arg\min_{\Theta} \sum_{i=1}^{n}\mathcal{L}_{\textrm{NLL}}\bigl(\mathcal{F}_{\Theta}(\mathcal{X}_i^{m_i}), \mat{y}_i\bigr).
\label{eq:training}
\end{equation}
As the training process progresses and the corresponding objective function is minimised, the SparseSense network uncovers highly discriminative embeddings for sparse segments that allow effective separation of classes in the activity space. 

Once the training procedure converges and the optimal network parameters $\Theta^*$ are learned from the training dataset, we adopt a maximum a posteriori (MAP) inference to promote the most probable activity category for any given set of sparse sensor readings; \textit{i.e.}, the highest scoring class in the softmax output of the network is chosen to be the final prediction.

%% file: Experiments.tex
\section{Experiments and Results}

\subsection{Datasets}

\begin{figure*}[t]
	\centering
	\includegraphics[width=\textwidth, keepaspectratio]{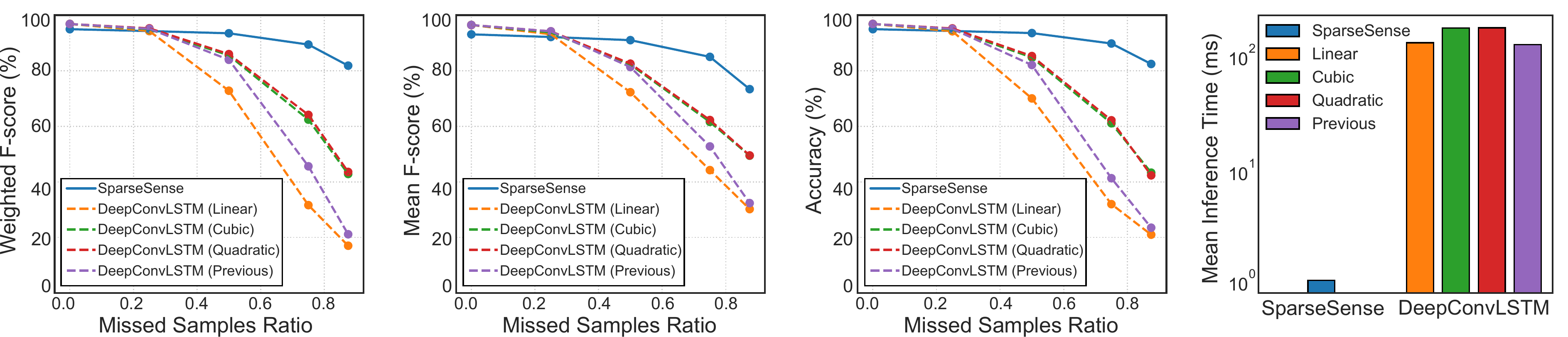}
	\caption{Activity recognition performance and computational complexity of our proposed \textit{SparseSense} framework for sparse data-stream classification against the state-of-the-art DeepConvLSTM HAR model. DeepConvLSTM is tailored for sensory segments of fixed size temporal context and thus, requires sparse segments be re-sampled through adoption of interpolation methodologies prior to inputting them.}
	\label{fig:drop}
\end{figure*}

To ground our study, we evaluate our proposed framework on two naturally sparse public datasets collected in clinical rooms with older people using a body-worn batteryless sensor intended for ambulatory monitoring in hospital settings. For further insights, we also present extensive empirical analysis of our approach on a HAR benchmark dataset with synthesized sparsification and provide comparisons against the state-of-the-art deep learning based HAR models.
\vspace{1mm}


\noindent\textbf{Clinical Room Datasets} \cite{clinical}:~The dataset is collected from  fourteen older volunteers, with a mean age of 78 years, performing a set of broadly scripted activities while wearing a $\textrm{W}^{2}\textrm{ISP}$ over their attire at the sternum level (see Fig. \ref{fig:workflow}). The $\textrm{W}^{2}\textrm{ISP}$ is a passive sensor-enabled RFID (Radio Frequency Identification) device that operates on harvested electromagnetic energy emitted from nearby RFID antennas to send data with an upper-bound sampling rate of 40~Hz. Data collection was carried out in two clinical rooms with two different antenna deployment configurations to power the sensor and capture data; resulting in \textit{Roomset1} and \textit{Roomset2} datasets. Each sensor observation in the obtained datasets records triaxial acceleration measurements as well as contextual information from the RFID platform indicating the antenna identifier and the strength of the received signal from the sensor. These recordings were manually annotated with \textit{lying on bed}, \textit{sitting on bed}, \textit{ambulating} and \textit{sitting on chair} to closely simulate hospitalized patients' actions. Consecutive samples in the sparse datastreams from \textit{Roomset1} and \textit{Roomset2} exhibit high mean time differences of $0.37$~s and $0.72$~s respectively.
\vspace{1mm}

\noindent\textbf{WISDM Benchmark Dataset} \cite{wisdm}:~ This dataset contains acceleration measurements from 36 volunteers collected through controlled, laboratory conditions while performing a specific set of activities. The sensing device used for data acquisition is an Android mobile phone with a constant sampling rate of 20 Hz and placed in the subjects' front pant's pocket. The sensor samples carry annotations from \textit{walking}, \textit{jogging}, \textit{climbing up stairs}, \textit{climbing down stairs}, \textit{sitting} and \textit{standing}. The collected dataset delivers high quality data and has frequently been used in HAR studies for benchmarking purposes. Accordingly, we find this dataset a suitable choice for thorough investigation of our SparseSense network under different levels of synthesized data sparsification.

\subsection{Experiment Setup}

In this study, we initially perform per-feature normalization to scale real-valued observation attributes to the $[0,1]$ interval. We consider a fixed temporal context $\delta t$ and obtain sensory partitions by sliding a window over the recorded data-streams. The acquired segments are assumed to reflect adequate information related to a wearer's current activity and are thus, assigned a categorical activity label based on the most observed sample annotation in the timespan of the sliding window. 

We implement the experiments in Pytorch \cite{pytorch} deep learning framework on a machine with a NVIDIA GeForce GTX 1060 GPU. The SparseSense deep human activity recognition model is trained in a fully-supervised fashion by back-propagating the gradients of the loss function in mini-batches of size 128; \textit{i.e.}, the network parameters are iteratively adjusted according to the RMSProp~\cite{rmsprop} update rule in order to minimise the negative log-likelihood loss using mini-batch gradient descent. The optimiser learning rate is initialised with $10^{-4}$, reduced by a factor of $0.1$ after 100 epochs, and the optimisation is ceased after 150 epochs. Further, a weight decay of $10^{-4}$ is imposed as $L_2$ penalty for regularisation. Following previous studies, we employ 7-fold stratified cross-validation on the datasets and preserve  activity class distributions across all folds. Each constructed fold is in turn utilized once for validation while the remaining six folds constitute the training data. 




\subsection{Baselines and Results} \label{results}

\noindent\textbf{Clinical Room Experiments.~} In Table \ref{tab:compare}, we report the mean F-measure ($\textrm{F-score}_{m}$) as the widely adopted evaluation metric and compare SparseSense with the highest performing activity recognition models previously studied for the naturally sparse clinical room datasets as well as the state-of-the-art deep learning based HAR models. Previous studies have explored shallow models including support vector machines (\textit{$\textrm{SVM}^{lin}$} and \textit{$\textrm{SVM}^{rbf}$}), and conditional random fields (\textit{$\textrm{CRF}$}) trained on top of hand-crafted features extracted from either raw or interpolated sparse segments. In addition, we investigate the effectiveness of \textit{Bi-LSTM} \cite{bilstm}, \textit{DeepCNN} and \textit{DeepConvLSTM} \cite{deepconvlstm} as solid deep learning baselines representing the state-of-the-art for HAR applications. 

Bi-LSTM leverages bidirectional LSTM recurrent layers to directly learn the temporal dependencies of samples within the sensory segments. Both DeepCNN and DeepConvLSTM adopt four layers of 1D convolutional filters along the temporal dimension of the fixed size segmented data to automatically extract feature representations. However, DeepCNN is then followed by two fully connected layers to aggregate the feature representations while DeepConvLSTM utilizes a two layered LSTM to model the temporal dynamics of feature activations prior to the final softmax layer. We refer interested readers to the original papers introducing the HAR models for further details and network specifications. Following \cite{interpolate}, for each baseline we explore progressively increasing window durations, \textit{i.e.} $\delta t\in\{2, 4, 8, 16\}$, adopt per-channel interpolation schemes (\textit{linear}, \textit{cubic}, \textit{quadratic} and \textit{previous}) to compensate for the missing acceleration data and report the highest achieving configurations in Table \ref{tab:compare}. In this regard, \textit{cubic} and \textit{quadratic} interpolation schemes respectively refer to a spline interpolation of second and third order, and the \textit{previous} scheme fills missed values with the previously received sensor readings.



From the outlined results, we observe that the SparseSense network outperforms all the baseline models with a large margin in the task of sparse data-stream classification. Notably, the baselines are: \textit{i}) well-engineered shallow models that require a large pool of domain expert hand-crafted features; and \textit{ii}) state-of-the-art deep learning HAR models that demand interpolation techniques to synthesize regular sensor sampling rates. In contrast, SparseSense seamlessly operates on sparse sets of sensory observations without requiring any extra interpolation efforts or manually designed features, and automatically extracts highly discriminative embeddings for the classification task in an end-to-end framework. 
\vspace{1mm}

\begin{figure}[t!]
	\centering
	\subfloat[DeepConvLSTM]{{\includegraphics[width=\columnwidth, keepaspectratio]{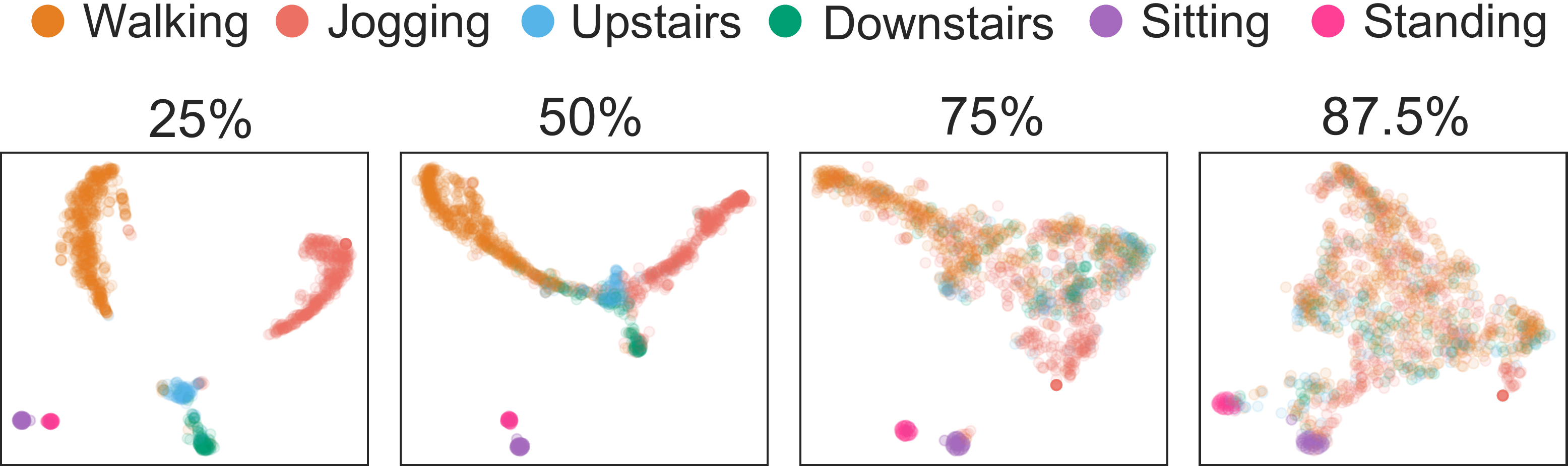} }}\\[-0.1ex]
	\subfloat[SparseSense]{{\includegraphics[width=\columnwidth, keepaspectratio]{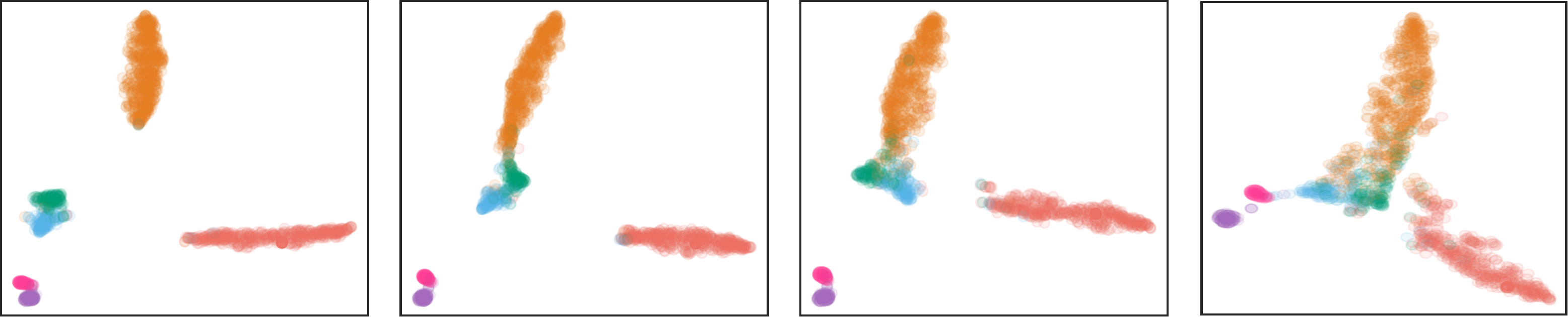} }}%
	\caption{A 2D visualization of the learned feature spaces for (a) DeepConvLSTM and (b) SparseSense under different data sparsity levels indicated by the percentage of artificially simulated missed readings. SparseSense learns robust embeddings that maintain cluster separation under significant imposed missing sample settings.} 
	\label{fig:embeddings}%
\end{figure}

\begin{figure}[t]
	\centering
	\includegraphics[width=\columnwidth, keepaspectratio]{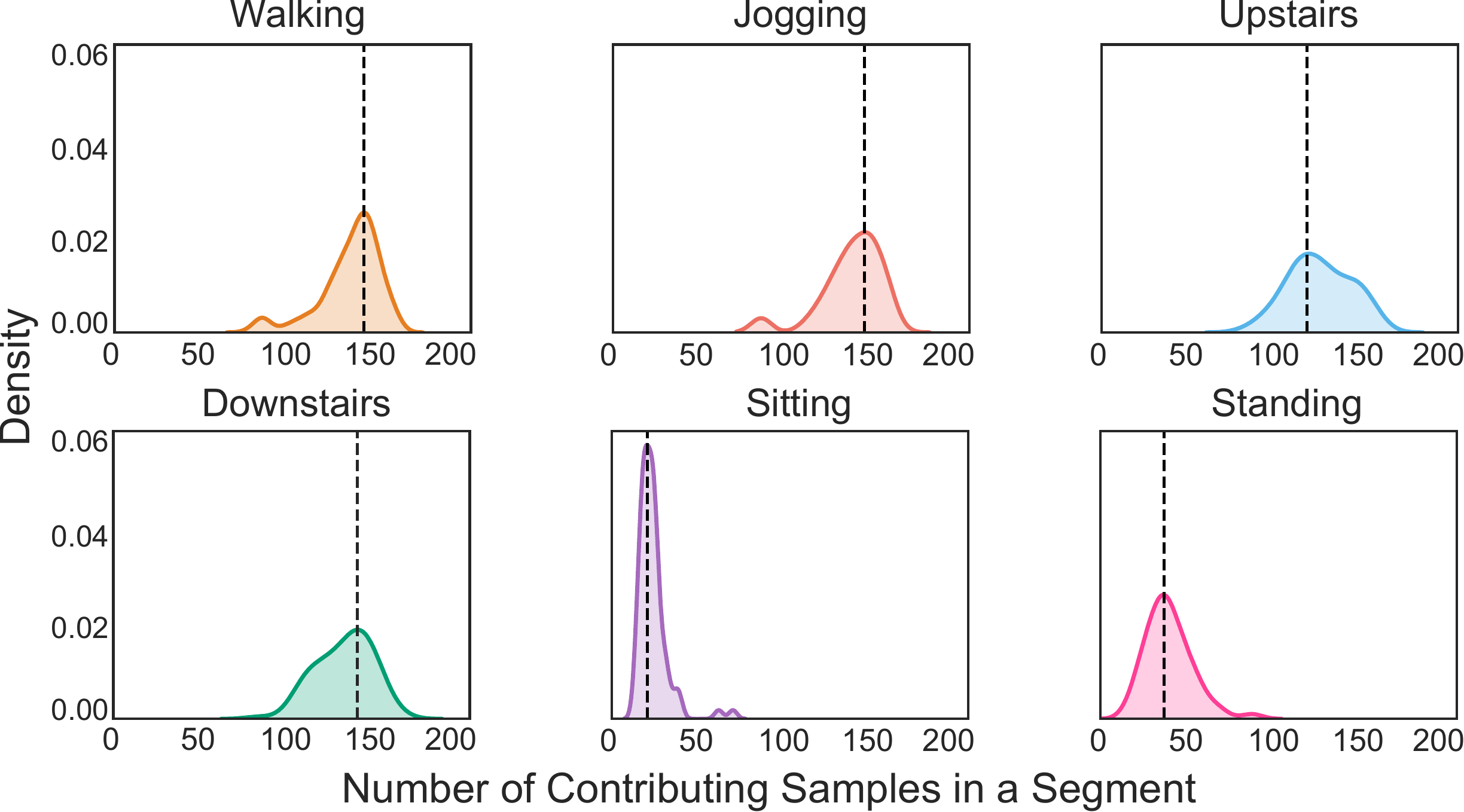}
	\caption{The density plots for the number of contributing samples that constitute the aggregated segment embeddings for each activity category of WISDM dataset.} 
	\label{fig:contrib}
\end{figure}


\noindent\textbf{WISDM Benchmark Experiments.\label{benchmarkexp}~} To provide additional insights onto the model's behavior, we conduct experiments on WISDM benchmark dataset and analyze the network's classification performance under different levels of synthesized data sparsification. Taking into account the superior performance of DeepConvLSTM among the baselines in Table \ref{tab:compare}, here we only present comparisons with this model. Following \cite{wisdm,alsheikh}, we partition the data-streams into fixed size sensory segments using a sliding window of 10 seconds duration (corresponding to 200 sensor readings) and train the HAR models on the acquired segmented data. Subsequently at test time, we drop sensor readings at random time-steps  in order to generate synthetic sparse segments. 
\vspace{1mm}

\noindent\textbf{Tolerance to Data Sparsity and Delays.}~In Fig. \ref{fig:drop}, the obtained evaluation measures  are plotted for both HAR models under different sparsification settings. When data segments are received in full, DeepConvLSTM performs better than SparseSense due to its ability in capturing temporal dependencies between consecutive sensor readings. However, as the data sparsity increases and the temporal correlation weakens, we observe a significant drop in classification performance of DeepConvLSTM. Notably, with large temporal gaps between sensor observations, interpolation techniques cannot produce good estimations of the missing samples and fail to recover the original acceleration measurements which in turn impacts the classification decisions of DeepConvLSTM. In contrast, not only does SparseSense achieve comparable classification results for completely received sensor data segments, but it also displays great robustness to data sparsity by making accurate decisions for incomplete segments of sensor data. In addition, we show in the bar plot the mean processing time required by the HAR models to make predictions on a mini-batch of 128 sensory segments. Clearly, our framework demonstrates a significant advantage over other HAR models for real-time activity recognition using sparse data-streams by removing the need for prior interpolation pre-processing.  
\vspace{1mm}

\noindent\textbf{SparseSense Model Behaviour.}~We visualize the high-dimensional feature space for both models in 2D space using t-distributed stochastic neighbor embedding (t-SNE) \cite{tsne} in Fig. \ref{fig:embeddings}. In the absence of significant data sparsity, the segment embeddings for each activity are clustered together while different activities are separated in the feature space for both models. However, while SparseSense is able to maintain this cluster separation for severely missed sample ratios and incomplete observation sets, DeepConvLSTM clearly struggles to discriminate between the interpolated segments. Technically, the symmetric max pooling operation in the aggregation layer of SparseSense incentivises our HAR model to summarise sensory segments using only the most informative sensor readings in the segment. 

In Fig. \ref{fig:contrib}, we provide density plots for the number of sensor readings that ultimately contribute to the aggregated segment embeddings for each activity category of the WISDM dataset. We observe that SparseSense intelligently summarizes the segments through discarding potentially redundant information in the neighboring samples when complete sensor data sets are presented to the network--see the density plots where the tails towards 200 contributing samples have a probability of zero. More interestingly, the network displays a clear distinction in its behavior towards learning embeddings for static activities (\textit{i.e.}, sitting and standing) as opposed to dynamic activities (\textit{i.e.}, walking, jogging and climbing stairs) by exploiting far fewer number of sensor observations in the window. This can be intuitively understood as static activities reflect signal patterns with small changes in sensor measurements of a timed window as compared with dynamic activities and thus, can be summarised with smaller number of observations.

%% file: Conclusions.tex
\section{Conclusions}
In this study, we present an end-to-end human activity recognition framework to learn directly from temporally sparse data-streams using set-based deep neural networks. In contrast to previous studies that rely on interpolation pre-processing to synthesise sensory partitions with fixed temporal context, our proposed \textit{SparseSense} network seamlessly operates on sparse segments with potentially varying number of sensor readings and delivers highly accurate predictions in the presence of missing sensor observations. Through extensive experiments on publicly available HAR datasets, we substantiate how our novel treatment for sparse data-stream classification results in recognition models that significantly outperform state-of-the-art deep learning based HAR models while incurring notably lower real-time prediction delays. 